# Text sampling strategies for predicting missing bibliographic links

F. V. Krasnov[a], I. S. Smaznevich[a]*, E. N. Baskakova[a]

[a] *NAUMEN R&D, 49A, Tatishcheva st., Yekaterinburg, 620028, Russian Federation*

*ismaznevich@naumen.ru

Krasnov Fedor Vladimirovich – Ph. D. in Engineering (Doctor of Technical Sciences). Expert at the Department of Semantic Systems, NAUMEN R&D. Research interests: intelligent text analysis. Author of over 70 scientific publications.
ORCID: 0000-0002-9881-7371
e-mail: fkrasnov@naumen.ru

Smaznevich Irina Sergeevna – Business analyst at the Department of Semantic Systems, NAUMEN R&D. Graduate of the Faculty of Computational Mathematics and Cybernetics, Lomonosov Moscow State University. Research interests: use of intelligent algorithms in applied information systems. The number of scientific publications – 8.
ORCID: 0000-0002-5996-4635
e-mail: ismaznevich@naumen.ru

Baskakova Elena Nikolaevna – Leading system analyst at the Department of Semantic Systems, NAUMEN R&D. Graduate of the Faculty of Informatics and Control Systems, Bauman Moscow State Technical University. Research interests: use of intelligent algorithms in applied information systems. The number of scientific publications – 5.
ORCID: 0000-0002-7071-8961
e-mail: enbaskakova@naumen.ru

Word count – 3199

# Text sampling strategies for predicting missing bibliographic links

The paper proposes various strategies for sampling text data when performing automatic sentence classification for the purpose of detecting missing bibliographic links. We construct samples based on sentences as semantic units of the text and add their immediate context which consists of several neighboring sentences. We examine a number of sampling strategies that differ in context size and position.

The experiment is carried out on the collection of STEM scientific papers. Including the context of sentences into samples improves the result of their classification. We automatically determine the optimal sampling strategy for a given text collection by implementing an ensemble voting when classifying the same data sampled in different ways. Sampling strategy taking into account the sentence context with hard voting procedure leads to the classification accuracy of 98% (F1-score). This method of detecting missing bibliographic links can be used in recommendation engines of applied intelligent information systems.

Keywords: text sampling, sampling strategy, citation analysis, bibliographic link prediction, sentence classification.

## 1. Introduction

Scientific research is impossible without correlating the results obtained with the work of other scientists. Other works should be mentioned by inserting bibliographic links in the article. Experts in scientometrics rationalize the need to establish such links between studies and formulate various citation theories.

The normative theory of citation, which draws on the principles of scientific ethics formulated by Merton (1973), assumes that references in scientific papers are made in order to indicate the works that are the basis for research or topically related, describe the research methods used and are necessary to discuss the results. According to the reflexive theory, links between scientific works indicate the state of science and help to create its formalized representation, e.g. maps of science (Akoev et al. 2014).

Thus, the beneficiary of scientific citation correctness is the entire scientific community, both researchers who create articles on their results and administrators who monitor achievements in various scientific fields. Mentioning relevant and remarkable results of other scientists is one of the basic requirements in the construction of scientific texts, in particular from the point of view of the editors of scientific journals. These requirements are noted in academic writing guidelines (Emerson et al. 2005; Gray et al. 2008; Pears and Shields 2019) and are confirmed in practice, for example, by the results of studies of publication activity in top-rated international journals (Arsyad et al. 2020).

Authors of scientific papers choose the sources for citation and positions for the links by themselves and at present, this process is not automated. In this work, we investigate the possibility of creating a recommendation algorithm that allows one to find missing bibliographic references in a scientific article, that is, to identify those text fragments where it is necessary to mention another research work. For this purpose, we estimate the probability of link presence in fragments of the text using a semi-supervised machine learning approach. The formal statement of the problem under consideration is the following: it is required to automatically find in the text of a scientific article those fragments (sentences) where the link is absent, but necessary, using a set of labeled fragments with and without links as training data.

The task of classifying text fragments in relation to the presence of the links in them is methodologically similar to the task of Sentiment Analysis, in which texts are automatically classified, mainly as positive and negative, according to their emotional characteristics. In addition to dividing fragments into positive and negative, the sentiment analysis approach is used to distinguish other classes, including citation significance detection (Aljuaid et al. 2021; Prester et al. 2021; Varanasi et al. 2021;

Färber and Ashwath 2019). The problem of identifying missing or unnecessary links in the text resembles sentiment analysis and the sought-for sentiment here is the author's need to confirm the formulated statement.

Another close line of research is Named Entity Recognition (NER) using prediction by classifiers. A similar problem is considered in (Fu et al. 2021), where NER problem is solved in the Span Prediction approach. NER can be performed in two stages: identifying fragments with a high probability of containing entities, and determination of the exact positions of these entities (Ziyadi et al. 2020; Li 2021). Some methods of NER also take into account the context of entities, both local and global, or external (Wang et al. 2021).

The task of sentence classification accounting for their nearest context has been discussed in a number of studies. Fiok et al. (2020) used contextualized embeddings created by language models, which high quality comes at the price of speed. Glazkova (2020) studied topical classification and showed that models taking context as input performed better than context-free models. In those works, context size is determined once based on some bias and may not be optimal for a certain text corpus.

The method introduced in this work also can be considered as kind of a resampling technique. Until now resampling has been used mainly for the purpose of balancing the class distribution in training datasets in order to improve the accuracy of class prediction, which is negatively affected by imbalanced data. Resampling methods are classified into three types, namely undersampling, oversampling and hybrid techniques. Undersampling eliminates some data of a majority class (see, e.g., Maya and Jayasudha 2017; Akkasi et al. 2017), while oversampling either replicates existing instances of a minority class or creates new ones (Luo et al. 2019; Li et al. 2018;

Chawla et al. 2002), and hybrid resampling techniques aim to combine benefits of both (Taha et al. 2021).

Local and global contexts are taken into account by modern neural network architectures for text analysis. Since text is a unidirectional list of terms, context is usually understood as some neighboring words before or after the term under consideration (Gallant 1991; Huang 2012). In convolutional neural networks, an increase in the size of context leads to a significant increase in the dimension of tensors and, consequently, in the number of parameters of a model, which requires an increase in the size of collections. In deep learning models known as transformers, context is explored by means of an attention mechanism, and the local context is combined with the broader context (BERT [Devlin et al. 2018], GPT-3 [Brown et al. 2020]).

It is important that all of the above algorithms do not take into account the natural structural units of texts (i.e. sentences and paragraphs) since these algorithms are adjusted to a certain size of the context, which is a fixed number of words, while the size of sentences and paragraphs varies.

**2. Methods**

The task of determining missing links is formalized as finding text fragments where the link is absent, but necessary, or, conversely, is present, but not needed.

We solve the problem using automatic binary classification with two classes namely positive and negative. For each fragment of a scientific article, our algorithm determines the probability of a bibliographic link in it. A collection of text documents is given such that each document consists of fragments. A fragment is a sequence of words (terms) of different lengths. Fragments can overlap each other and vary in size. Each fragment is a sample and is labeled as one of the two possible classes with class labels: positive or negative. The class label corresponds to whether or not the given fragment contains a bibliographic link. The task of this study is to find a strategy for the construction samples of fragments, which gives the highest accuracy in determining the labels of the class by a certain classifier.

The hypothesis of our study is the following: text sampling strategies that take into account the context increase the accuracy of sentence classification used to predict missing bibliographic links in scientific articles.

We suggest that a positive sample consists of a bibliographic link surrounded by its context from the original text, and a negative sample is a fragment with no bibliographic link in it. In order to avoid duplication of samples, we consider a sentence with two or more links only once. The context of the link is limited to the sentence containing it, or the context is extended and it includes neighboring sentences as well.

The best option is when the boundaries of a link context coincide with the boundaries of the complete author's statement to which this link belongs. In this case, a semantic unit of text can be either one or several sentences, which makes it difficult to set the size of the context. Nevertheless, to approach the specified goal in the proposed algorithm, as a context we consider a fragment which size is determined by the number of sentences, and not words, unlike neural network algorithms. Thus, in our algorithm context is formed on the basis of natural structural units of text.

The feature space is constructed automatically based on vocabulary statistics within the Bag-of-Words model (BoW). The vocabulary of the model includes words and all the original punctuation marks and typographical symbols. As additional features, we consider named entities.

*Algorithm*

The algorithm consists of the following stages.

*1. Text preprocessing*

- Text cleaning: removal of service characters (tabulation, line feed, etc.), words (journal titles, ISBN, etc.), and sections (funding, reference);
- Tokenization:
    - splitting text into sentences;
    - terms normalization.

*2. Data labeling*

- For each document (journal article), beginning and end marks are added;
- For each sentence:
    - if the sentence contains a bibliographic link (citation marker), it is labeled as belonging to the class "With links";
    - if there is no citation marker in the sentence, it gets "Without links" class label.
    - After labeling sentences, citation markers are removed.

*3. Named Entities processing*

- Detection of named entities in the text;
- Replacement of named entities with special marks.

*4. Construction of samples*

Samples are constructed in different ways depending on the class (positive or negative):

- To form a positive sample, we take a sentence "With links" and add *n* previous sentences and *m* subsequent sentences, all sentences are taken in the original order.

- A negative sample is constructed of *k* sentences "Without links" in a row in the original text (adjacent sentences), where *k = n+1+m*.

The visualization of the contents of the samples is shown in Figure 1.

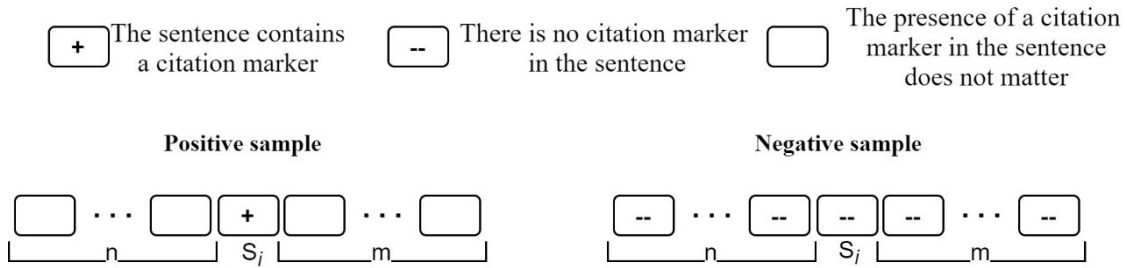

Figure 1. Construction of samples.

*5. Classification of samples*

- Balancing the class distribution in the training set is done by random undersampling.
- For each sample, a vector model is built using count vectorizer as the fastest and most computationally effective text representation.
- Vectorized set of samples is processed using a classifier.

*6. Optimal sampling strategy determination*

Further, an ensemble method is used to automatically determine the optimal sampling strategy. We give the same data sampled in different ways to the estimators of the same type and implement a voting procedure.

The flowchart of the whole algorithm is shown in Figure 2. Each $BoW_j$ corresponds to one sampling strategy, and for each sampling strategy, we run its own estimator. All the estimators implement the same classification method but take different types of samples as input data.

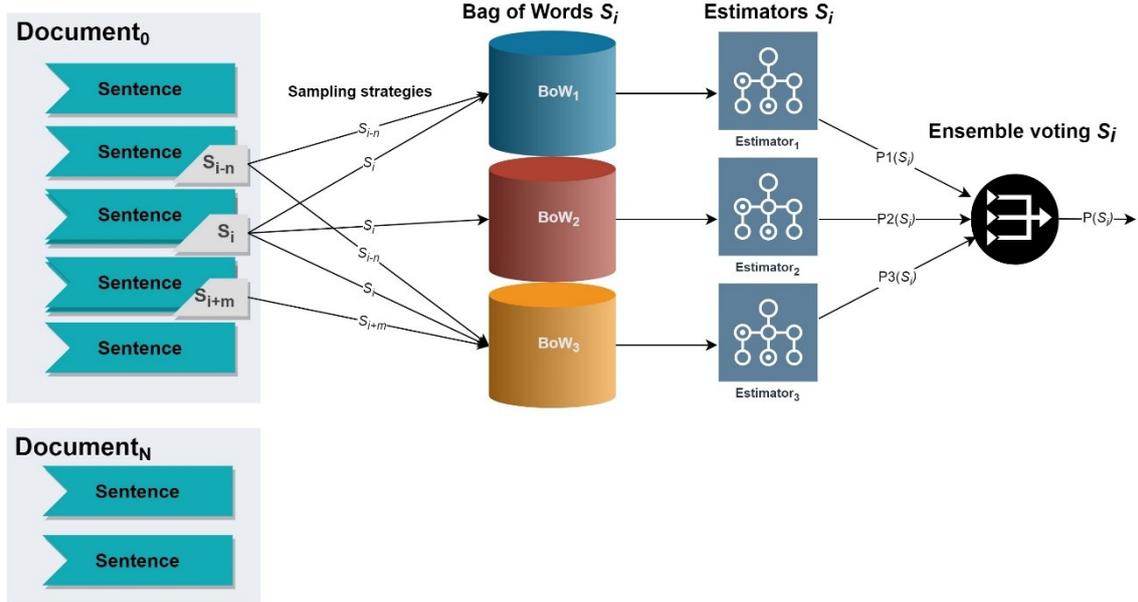

Figure 2. The algorithm flowchart for classifying sentences into the classes "With links" and "Without links" using various sampling strategies (for *n*=*m*=1).

## 3. Experiment

To test the hypothesis experimentally, we took the dataset of STEM journal articles, collected from scientific repository arXiv.org by Cohan et al. (2018). Documents of this dataset contain only texts, while figures and tables are removed. Math formulas and citation markers are replaced with special tokens @xmath<number> and @xcite. Documents contain only the sections up to the conclusion section and all sections after the conclusion are removed.

The size of the dataset is the following: the number of documents – 215K, the average document length – 4938 words, the average summary length – 220 words.

The files are in jsonlines format where each line is a json object corresponding to one scientific article. Each line contains an abstract, sections and a body of the article, and all these texts are sentence tokenized.

In our experiment we consider sentences that are more than 30 words long. With this restriction we got the set of 458774 sentences in total.

Sentences containing citation markers @xcite are assigned to the positive class ("With links"), and after that citation markers are removed. Sentences without citation markers are labeled as negative ("Without links"). The ratio of classes is: 24% – positive sentences, 76% – negative sentences. This is assumed as sampling strategy #0, and the classification result on data sampled that way is considered as a baseline: classification accuracy with sampling strategy #0 measured by F1-score is 0.7866.

After establishing the baseline, we test various strategies of data sampling in order to improve the classification accuracy. The main idea of sampling is to take into account some context of the sentences with a link. Different strategies of sampling assume various directions, positions, and sizes of the context, determined by a number of surrounding sentences. In different sampling strategies, each sentence [$i$] is included in different types (variants) of samples.

In the experiment we test 10 strategies with the following parameters: $n$: [0, 1, 2 3 4 5], $m$: [0, 1, 2, 3, 4], $k$: [1, 3]. All the sample types corresponding to the chosen sampling strategies are presented in Table 1.

Table 1. Sampling strategies tested in the experiment.

| # | Sampling strategy (sample constructing algorithm) | Number of sentences per sample | |
|---|---|---|---|
| | | Positive | Negative |
| 0 | Sentence [$i$] | 2640 | 12352 |
| 1 | Sentences [$i$: $i$ +2] | 2640 | 10639 |
| 2 | Sentences [$i$ -1: $i$+1] | 2640 | 10639 |
| 3 | Sentences [$i$ -1: $i$+2] | 2640 | 9376 |
| 4 | Sentences [$i$ -2: $i$+2] | 2640 | 8384 |
| 5 | Sentences [$i$ -3: $i$+2] | 2640 | 7574 |
| 6 | Sentences [$i$ -3: $i$+3] | 2640 | 6880 |
| 7 | Sentences [$i$ -4: $i$+3] | 2640 | 6287 |
| 8 | Sentences [$i$ -4: $i$+4] | 2640 | 5776 |
| 9 | Sentences [$i$ -5: $i$+4] | 2640 | 5326 |

The distribution of the length (number of words) in positive and negative samples of different types is shown in Figure 3.

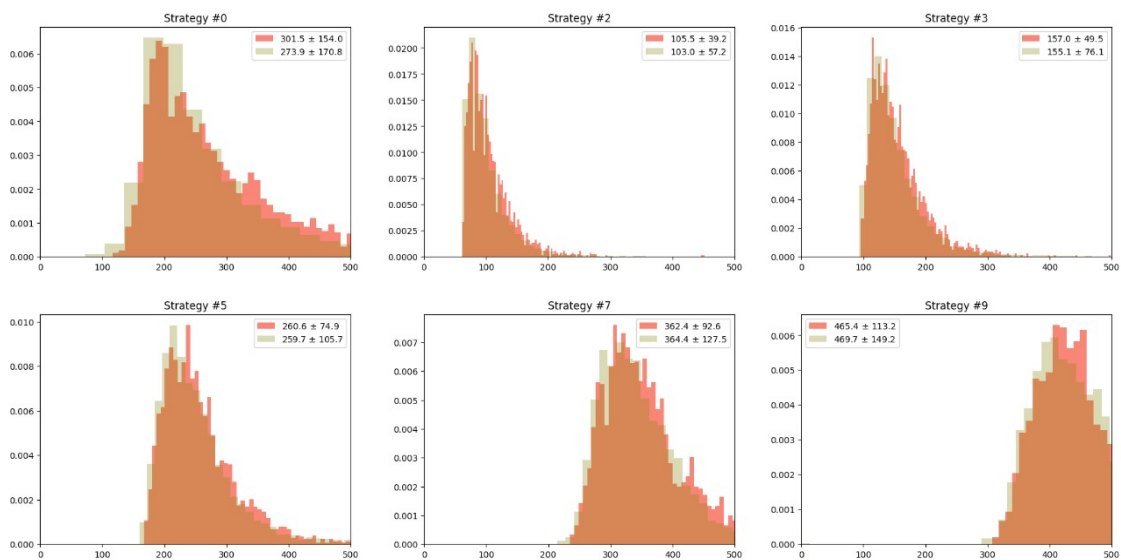

Figure 3. The distribution of the length as a number of words in positive and negative samples of different types ('salmon' color refers to the positive class, 'olive' refers to the negative one).

After equalizing classes by random undersampling the data is divided into training and test sets with the proportion parameter test_size=0.33.

Vector representation is build using CountVectorizer method of the Scikit-learn library. The vocabulary includes unigrams and bigrams and is reduced by frequency with the parameters min_df=3, max_df=0.7

For classification we use a neural network multilayer perceptron (the MLPClassifier method of the Scikit-learn library). Classification performance depending on the sampling strategy used is presented in Table 2.

Table 2. The result of sentence classification by the MLPClassifier depending on the sampling strategy (weighed average).

| Sampling strategy | F1 | Precision | Recall |
|---|---|---|---|
| 0 | 0.7866 | 0.7866 | 0.7866 |
| 1 | 0.8882 | 0.8881 | 0.8881 |
| 2 | 0.8884 | 0.8881 | 0.8881 |
| 3 | 0.9214 | 0.9214 | 0.9214 |
| 4 | 0.9444 | 0.9443 | 0.9443 |
| 5 | 0.9410 | 0.9409 | 0.9409 |
| 6 | 0.9601 | 0.9598 | 0.9598 |
| 7 | **0.9640** | **0.9639** | **0.9639** |
| 8 | 0.9593 | 0.9593 | 0.9593 |
| 9 | 0.9581 | 0.9581 | 0.9581 |

It can be seen from the table that the best result is achieved using the sampling strategy #7. A further increase in the number of sentences per sample does not significantly improve the accuracy, since it tends to the asymptote.

For each sentence we compare the result of classification obtained with all the sampling strategies and further improve it by voting procedure, both soft and hard. For soft voting, we set the threshold value 0.5 for the mean predicted probability. For hard voting, we summarize all the predicted probabilities and compare the sum with the threshold value 3. We test different numbers of estimators. To form combinations of sampling strategies we start from the group of estimators #7, #8, and #9 and then add estimators one by one in reverse order. The voting results depending on the number of estimators considered are shown in Figure 4.

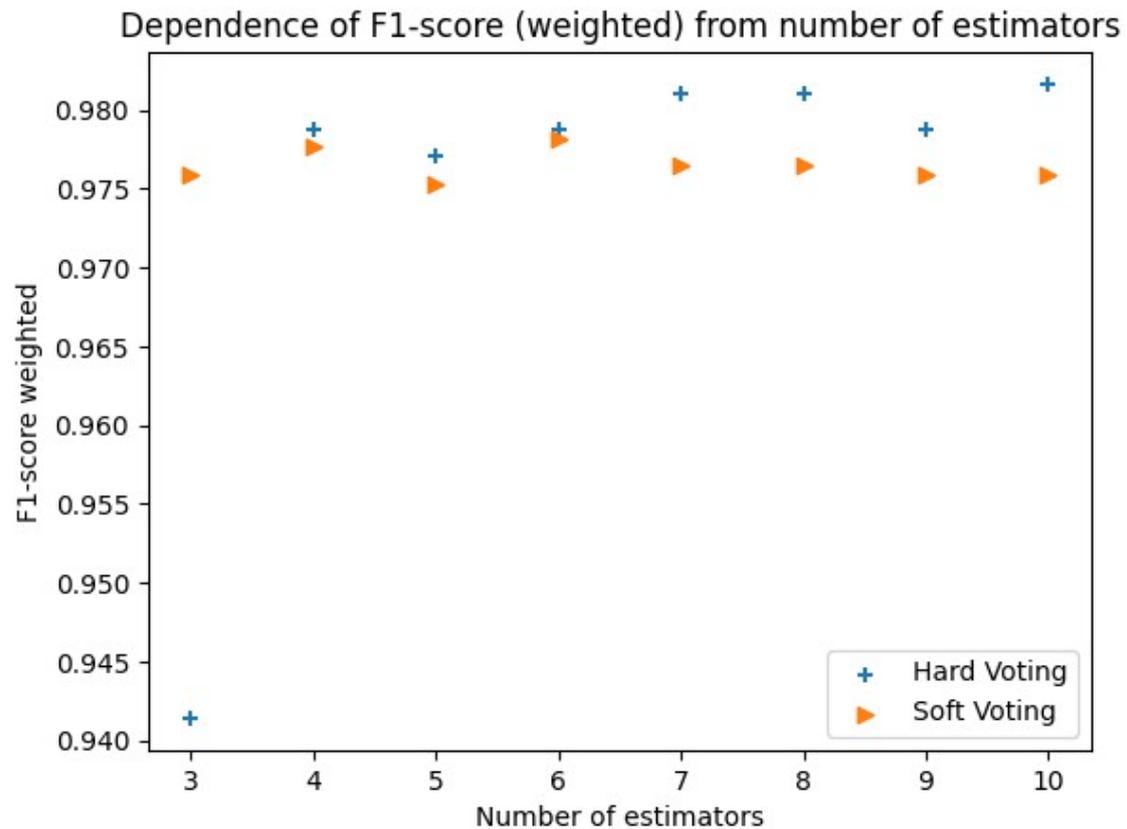

Figure 4. The result of classification with hard and soft voting depending on the number of estimators included.

The graph demonstrates that the voting procedure further improves classification performance and increases F1-score by 1,5%. With all the combinations of estimators tested the result is consistently high, but the best result (>98%) is achieved with hard voting of 7,8 or 10 estimators classifying long samples.

## 4. Result

The formulated research hypothesis has been confirmed experimentally. We have shown that the choice of the sampling strategy affects the result of text classification.

The baseline is established with sampling strategy #0. In this case, the classification performance measured using the F1-score is only 79 %, which is not sufficient for practical use in industrial information systems.

The improvement is achieved due to the data sampling strategy which assumes automatic determination of the optimal sample type. That is provided by applying the voting procedure to the decisions made by different estimators. The proposed algorithm shows 98% accuracy (F1-score), which is comparable to the state-of-the-art results for NER using automatic classification and other text classification tasks. It is important that the proposed algorithm provides high accuracy but doesn't require huge computational resources to be implemented.

## 5. Conclusion

The paper proposes a new method of determining the probability of a bibliographic link in fragments of a scientific article. The approach assumes sentence classification with ensemble voting, in which different data sampling strategies correspond to estimators implementing the same classification method. The problem statement made by the authors is close to well-studied areas NER and sentiment analysis but is new from the real application point of view.

The main innovation of the proposed method is finding the link context that maximally affects the probability of detecting a missing bibliographic link in a sentence. In the proposed algorithm, the best size and position of context are determined automatically. The size is based on the boundaries of semantic units of the text and is measured by the number of sentences, not words, thus we utilize the fact that a sentence is a more semantically capacious (meaningful) unit than a word. Most existing text classification methods do not assume fragment context as significantly important, but this study shows the critical importance of taking it into account. The considerable impact of the context on the classification performance demonstrates that semantics related to a bibliographic link can be localized in fragments of different lengths.

The accuracy of the proposed algorithm reaches 98% (F1-score). It is important to note the high computational efficiency of the described method in comparison with convolutional artificial neural networks. This advantage is achieved due to the bigger size of samples. The investigated approach to text analysis expands the principle of the attention mechanism aimed at training a language model to understand the impact of global and local context. Automatic determination of the context boundaries correlates with the idea of automatic selection of significant features in artificial neural networks.

The proposed method can be used in recommendation engines of applied intelligent information systems, including assistance for constructing documents and composing texts with probable links to other documents, or help in checking the document correctness. Such functions are useful in many fields e.g. science, law, or journalism, where documents contain statements that should be confirmed by references to legal acts or other sources.

In accordance with the company's policy, we do not publish the source code.